\pdfoutput=1
\documentclass[10pt,twocolumn,letterpaper]{article}

\usepackage{iccv}
\usepackage{times}
\usepackage{epsfig}
\usepackage{graphicx}
\usepackage{amsmath}
\usepackage{amssymb}

\usepackage{booktabs}
\usepackage{algorithm, algorithmic}
\usepackage{multirow}
\usepackage{multicol}
\usepackage{bbding}
\usepackage[table]{xcolor}
\usepackage{enumitem}

\usepackage{enumitem}
\usepackage{empheq}
\usepackage[most]{tcolorbox}

\usepackage[breaklinks=true,bookmarks=false]{hyperref}

\iccvfinalcopy 

\def\iccvPaperID{****} 

\ificcvfinal\fi

\begin{document}

\title{DLIP: Distilling Language-Image Pre-training}

\author{Huafeng Kuang\textsuperscript{1*}, \text{ } Jie Wu\textsuperscript{2}, \text{ } Xiawu Zheng\textsuperscript{3}, \text{ } Ming Li\textsuperscript{2}, \text{ } Xuefeng Xiao\textsuperscript{2}, \\
Rui Wang\textsuperscript{2}, \text{ } Min Zheng\textsuperscript{2}, \text{ } Rongrong Ji\textsuperscript{1,3} \\
$^{1}$ Xiamen University, \text{ } 
$^{2}$ Bytedance Inc, \text{ } 
$^{3}$ Peng Cheng Laboratory \\
{\tt\small skykuang@stu.xmu.edu.cn},
{\tt\small \{wujie.10, liming.ai, xiaoxuefeng.ailab\}@bytedance.com}, \\
{\tt\small \{ruiwang.rw, zhengmin.666\}@bytedance.com}, \\
{\tt\small zhengxw01@pcl.ac.cn}, 
{\tt\small rrji@xmu.edu.cn}
}

\maketitle
\ificcvfinal\thispagestyle{empty}\fi

\begin{abstract}
    Vision-Language Pre-training (VLP) shows remarkable progress with the assistance of extremely heavy parameters, which challenges deployment in real applications.
    Knowledge distillation is well recognized as the essential procedure in model compression. However, existing knowledge distillation techniques lack an in-depth investigation and analysis of VLP, 
    and practical guidelines for VLP-oriented distillation are still not yet explored. 
    In this paper, we present DLIP, a simple yet efficient Distilling Language-Image Pre-training framework, through which we investigate how to distil a light VLP model. 
    Specifically, we dissect the model distillation from multiple dimensions, such as the architecture characteristics of different modules and the information transfer of different modalities.
    We conduct comprehensive experiments and provide insights on distilling a light but performant VLP model.
    Experimental results reveal that DLIP can achieve a state-of-the-art accuracy/efficiency trade-off across diverse cross-modal tasks, e.g., image-text retrieval, image captioning and visual question answering.
    For example, DLIP compresses BLIP by \textbf{1.9$\times$}, from 213M to 108M parameters, while achieving comparable or better performance.
    Furthermore, DLIP succeeds in retaining more than \textbf{95\%} of the performance with \textbf{22.4\%} parameters and \textbf{24.8\%} FLOPs compared to the teacher model and accelerates inference speed by \textbf{2.7$\times$}.
\end{abstract}

\renewcommand{\thefootnote}{\fnsymbol{footnote}}

\footnotetext[1]{work done during internship at
Bytedance Inc.}

\section{Introduction}
\label{sec:intro}
Thanks to the emergence of foundation models, the large language and vision models are integrated to acquire the multimodal ability, and witness prevailing success on a wide range of multimodal downstream tasks, such as image-text retrieval \cite{plummer2015flickr30k}, image caption \cite{you2016image} and visual question answering (VQA) \cite{antol2015vqa}.
However, whether it is the Vision-Language Pre-training (VLP) \cite{li2021align,radford2021learning,li2022blip,yu2022coca} or the Multimodal Large Language Model (MLLM) \cite{li2023blip,liu2023visual,peng2023instruction,ye2023mplug} usually resort to a cumbersome model with hundreds of millions of parameters for inference.
It is arduous to popularize these models that require tremendous computational costs, which becomes a critical bottleneck when such models are required to deploy on resource-constrained server platforms or even more lightweight mobile devices.
To this end, building a light multimodal model to reduce the model parameters and computational overhead is crucial and practical in real-scene.

\begin{figure}[t]
  \centering
  \centerline{\includegraphics[width=\columnwidth]{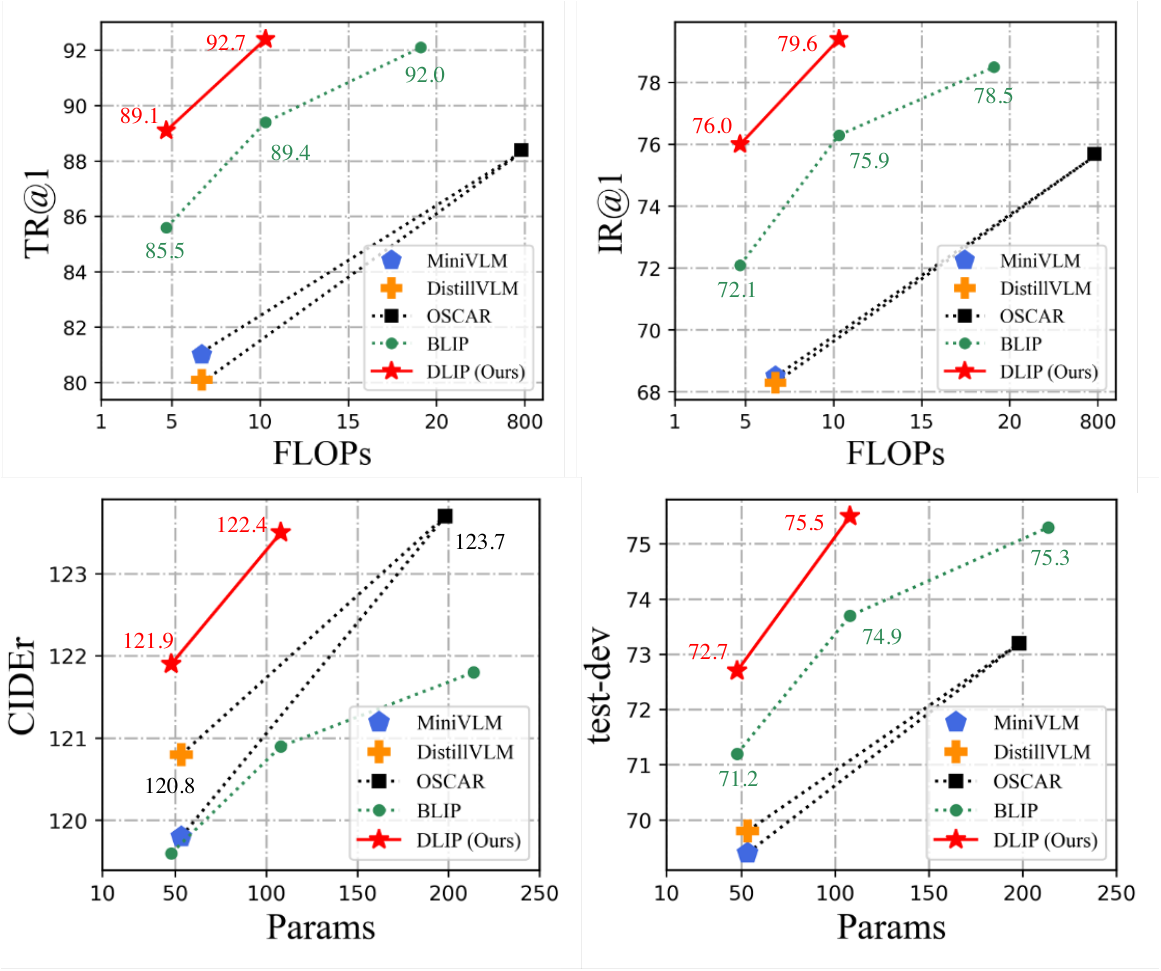}}
  \caption{ The trade-off between the performances (text retrieval
(TR) and image retrieval (IR) on Flickr30k, CIDEr of COCO caption and the test-dev score of VQAv2) and the FLOPs/Parameters. Comparisons between DLIP and existing methods. All models are trained on 4M training data. 
\textbf{Viewed in color}}
\label{trades}
  \vskip -0.2in
\end{figure}

Recent studies have demonstrated that there is redundancy in the deep learning model \cite{li2016pruning, molchanov2016pruning, he2017channel}. 
Therefore, a series of model compression technologies have been proposed to reduce the model size and accelerate model inference. 
Typical approaches to facilitate the model compression are based on knowledge distillation (KD) \cite{hinton2015distilling, polino2018model, sun2019patient, sanh2019distilbert}.
KD aims to transfer the knowledge learned in a large teacher model to a small student model,  which is trained to mimic the informative concepts (\emph{e.g., output logit, features maps or hidden representations}) of the teacher model. 
For example, DistBERT \cite{sanh2019distilbert} reduces the size of the BERT-base model by 40\% by closing the hidden representation of the teacher and the student model.
MiniLM \cite{wang2020minilm} further highlights the importance of minimizing the self-attention distributions across the teacher and student model. 
However, previous distillation algorithms fail to consider the impact of the complex architecture of vision-language models and the effects of multimodal information transfers.

\begin{figure*}[t]
      \begin{center}
      \centerline{\includegraphics[width=170mm,height=75mm]
      {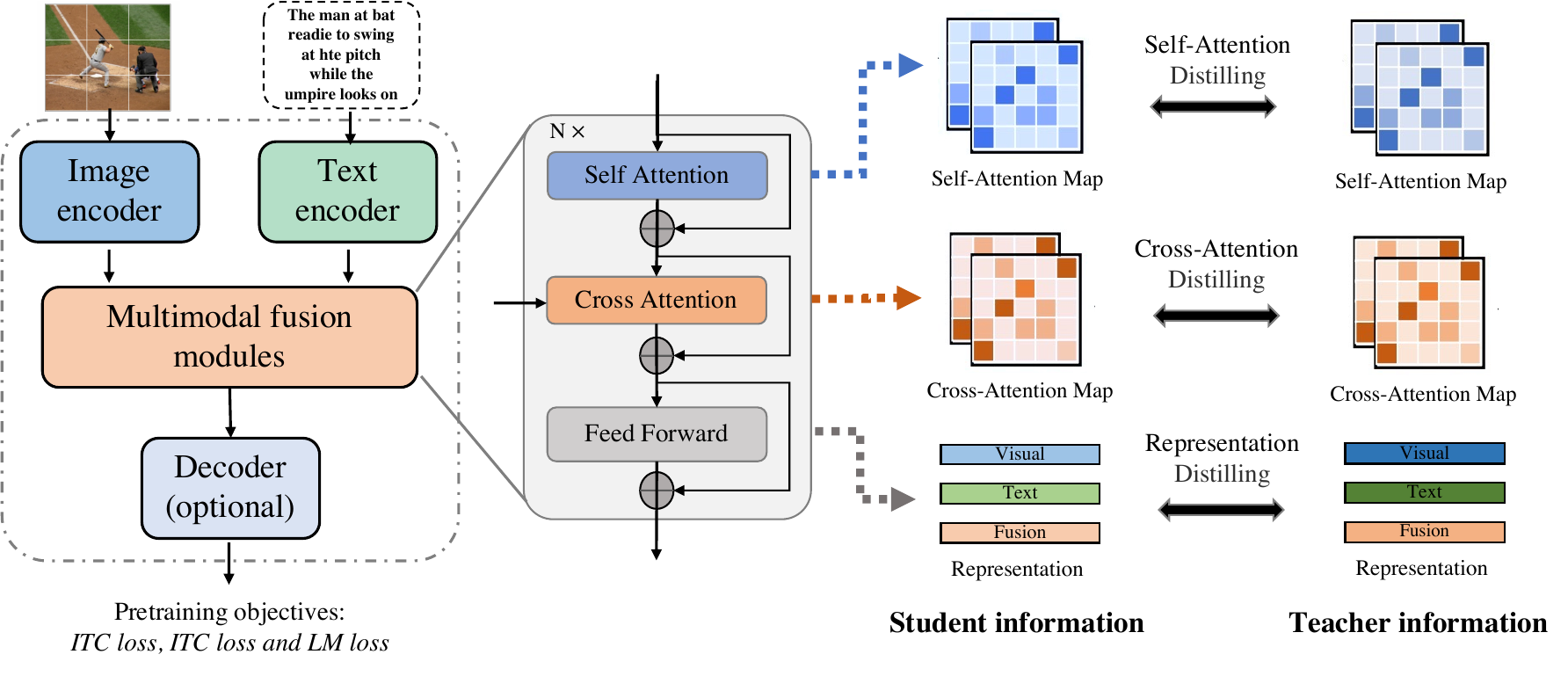}}
      \caption{
      The architecture of most VLP models usually consists of four modules, including an image encoder, text encoder, multimodal fusion module and decoder. In our DLIP framework, the student model is trained with two kinds of objectives: 1) Pre-training objectives, including image-text contrastive (ITC) loss, image-text matching (ITM) loss and language modelling (LM) loss; 2) Distilling objectives, including attention-based (\emph{e.g.,} self-attention and cross-attention) transfer loss and representation-based (\emph{e.g.,} visual representation, text representation and fusion representation) transfer loss.
      }
      \vspace{-3.0em}
      \label{framework}
      \end{center}
  \end{figure*}

For fully transformer-based VLP models \cite{kim2021vilt, li2022blip, dou2022empirical},
we observe that distil it confronts two fundamental challenges: 1) the architecture of VLP models usually contains multiple modules, including image encoders (\emph{e.g.,} ViT \cite{radford2021learning}), text encoders (\emph{e.g.,} BERT \cite{vaswani2017attention}), multimodal fusion modules or task decoders, as shown in Figure \ref{framework}.
Therefore, it is a non-trivial task to determine which modules could be distilled. 
2) Compared to unimodal (only vision or language) distillation, 
the VLP model involves a variety of information transfer, including unimodal information (\emph{e.g.,} visual information) and multimodal fused information (\emph{e.g.,} Visual-linguistic information).
Therefore, it is essential to investigate the impact of different information transfers on downstream tasks in distillation.

To address the abovementioned challenges, we conduct a series of controlled experiments to investigate in-depth and analyse the VLP models. We ablate the role of different module compression and pursue minimal changes made for pinpointing the impact of different modal information transfers in VLP distillation.
Through extensive analyses, we summarize our findings as follows:
\begin{itemize}[itemsep=5pt,topsep=5pt,parsep=0pt]
    \item  In the dimension of modules, image and text encoders are equally important in model compression. Moreover, the large fusion module is unnecessary, and moderate fusion layers are beneficial and efficient. 
    \item For the VLP model with a decoder, the decoder requires separate distillation to improve the performance of decoder-based downstream tasks.
    \item  In the dimension of modalities, the representation information is better than the attention information for distillation. Furthermore, multimodal information is more efficient than unimodal information. 
    \item  Initialization with pre-trained models is important for the visual encoder but has little impact on the text encoder. 
\end{itemize}

Based on the insights and other useful tricks detailed in later sections, we present Distilling Language-Image Pretraining (DLIP), a simple yet efficient VLP distillation framework for facilitating the training of smaller, faster and lighter VLP models, as shown in Figure \ref{framework}.
Perhaps surprisingly, without a complicated algorithm design, 
our DLIP shows very strong performance in efficiency and effectiveness and outperforms state-of-the-art baselines in different parameter sizes on broad multimodal tasks. 
In particular, DLIP compresses BLIP by \textbf{1.9$\times$}, from 213M to 108M parameters, while achieving comparable or better performance.
Furthermore, DLIP succeeds in retaining more than \textbf{95\%} of the performance with \textbf{22.4\%} parameters and 24.8\% FLOPs compared to the teacher model and accelerates inference speed by \textbf{2.7$\times$}.

\section{Related Work}
\subsection{Vision-Language Pre-training}
Pre-training and then fine-tuning have become a new paradigm for many deep learning tasks. 
Following the prominent progress in the pre-training models in natural language process \cite{vaswani2017attention, devlin2018bert, brown2020language, liu2019roberta, sanh2019distilbert, gururangan2020don} and computer vision \cite{he2022masked, he2020momentum, chen2020simple, chen2021empirical, bao2021beit, wei2022masked}, visual-linguistic pre-training (VLP) models have achieved great success on a number of downstream VL tasks, which pre-train the model on large-scale image-text pairs collected from the web since the prohibitive expense of acquiring human-annotated texts.
On the other hand, the great success of large language models (LLMs) \cite{brown2020language,ouyang2022training,hoffmann2022training,chung2022scaling} has
motivated the advancement of vision and multimodal in terms of architecture design and learning paradigm, therefore, multimodal large language models (MLLM) \cite{li2023blip,liu2023visual,peng2023instruction,liu2023aligning, ye2023mplug, openai2023gpt} have successfully played a role as a general-purpose interface across a wide range of vision-language multimodal tasks.
For instance, GPT-4 \cite{openai2023gpt} demonstrates impressive multimodal understanding and reasoning abilities, accepting image and text inputs while responding in text.

However, no matter whether VLP or MLLM, they are based on transformer \cite{dosovitskiy2020image}, and the models usually consist of hundreds of millions of parameters, which brings challenges for online serving and deployment in real-world applications due to latency and capacity constraints. 
Recently, some work has attempted to reduce the number of parameters of the model to improve the deployment and inference cost of the model. For example, MiniVLM \cite{wang2020minivlm} uses a lightweight visual feature extractor and smaller transformer to reduce the model size and maintain good accuracy on VL tasks.

\subsection{Knowledge Distillation}
Knowledge distillation (KD) \cite{hinton2015distilling, zagoruyko2016paying, park2019relational, beyer2022knowledge} is one effective method for transferring the knowledge in a larger teacher model $ h_t $ to a smaller student model $h_s$. The student model is trained to mimic the behaviour of the teacher network. The objective of KD can be formulated as follows:
\begin{equation}
  \min_{{\theta}_{S}} \mathbb{E} \bigg [ \mathcal{L}_{task} \big(h_{s}(x) , y \big) + \alpha \cdot L_{Dist} \big(h_{s}(x ), h_{t}(x )\big) \bigg ],
\end{equation}
where $\mathcal{L}_{task}$ is the task-specific loss function for the target task, and $ \mathcal{L}_{Dist} $ is a loss function that penalizes the difference between the teacher and the student. 
Several works leverage knowledge distillation to improve the performance of student models or compressing student models \cite{zagoruyko2016paying, li2021align, fang2021compressing, sanh2019distilbert, yang2022vitkd}.
ViTKD \cite{yang2022vitkd} explores the way of feature-based distillation for vision transformers.
DistillBERT \cite{sanh2019distilbert} leverages knowledge distillation to train a small BERT by mimicking the teacher’s output probability of masked language prediction and the embedding features.
In addition, \cite{zagoruyko2016paying, jiao2019tinybert, sun2020mobilebert, fang2021compressing} leverage attention-based distillation to compress the model, which forces the student to mimic the attention maps of a powerful teacher model to improve the performance of the student model significantly. 

In this work, we systematically explore the importance of different modules and the impacts of different information in distillation to  build a light multimodal model.

\section{The DLIP Framework}
This section introduces the DLIP framework, including the model architecture designs, pre-training objectives, distilling objectives and our default settings.

  \begin{table*}[t]
    \caption{Variants of vision encoder and text encoder architecture. The only parameters that vary across models for the vision encoder are the embedding dimension and the number of heads. For the text encoder, the parameters that vary across models are the embedding dimension and the number of layers. Smaller models have a lower parameter count and a faster inference. The FLOPs and inference are measured for images at resolution 224$\times$224 and text at length 30, respectively. 
    The inference is tested on Nvidia Tesla V100 GPU.}
    \label{model_cfg}
        \begin{small}
    \begin{center}
    \renewcommand\arraystretch{1.2}
    \setlength\tabcolsep{4.5pt}
    \begin{tabular}{c|ccccccc}
        \toprule
        \multirow{2}{1.5cm}{\centering Model}  & \multirow{2}{2.0cm}{ \centering Embedding dimension} & \multirow{2}{1.0cm}{ \centering \# Head} & \multirow{2}{1.5cm}{\centering \# Layers} & \multirow{2}{1.5cm}{\centering Params} & \multirow{2}{1.5cm}{\centering FLOPs} & \multirow{2}{1.5cm}{\centering Inference} & \multirow{2}{2.5cm}{\centering Input resolution/length} \\
        &  &  &  &  &  &  &  \\
        \toprule
        ViT-Base             & 768 & 12 & 12 & 85.6M (1.0$\times$) & 16.9G & 34.9ms & 224 \\
        ViT-Middle             & 576 & 9 & 12 & 48.3M (0.56$\times$) & 9.51G & 22.1ms & 224 \\
        ViT-Small                 & 384 & 6 & 12 & 21.6M (0.25$\times$) & 4.25G & 11.4ms & 224  \\
        ViT-Tiny                      & 192 & 3 & 12 & 5.48M (0.06$\times$) & 1.08G & 4.70ms & 224 \\
        \midrule
        BERT-Base                  & 768 & 12 & 12 & 137.3M (1.0$\times$) & 2.25G & 6.39ms & 30 \\
        BERT-Middle                  & 576 & 12 & 8 & 59.6M (0.43$\times$) & 0.88G & 4.03ms & 30 \\
        BERT-Small                  & 384 & 12 & 6 & 26.1M (0.20$\times$) & 0.32G & 3.49ms & 30\\
        BERT-Tiny               & 192 & 12 & 4 & 9.61M (0.07$\times$) & 0.10G & 2.21ms & 30 \\
        \bottomrule 
    \end{tabular}
    \end{center}
     \end{small}
    \vskip -0.2in
  \end{table*}

\subsection{Model Architecture}

A fully transformer-based VLP model usually consists of several important modules: a vision encoder, a text encoder, multimodal fusion modules and a task encoder, as shown in Figure \ref{framework}. Given an image-text pair $ < I, S> $, a VLP model first extracts both visual representation $V=\{ v_1, \dots, v_n \}$ and text representation $ T =\{ t_1 \dots t_m \} $ via a vision encoder and a text encoder, where $n$ and $m$ are the numbers of layers of encoders. Then the visual and text representations are fed into a multimodal fusion module to produce cross-modal representations, which are then optionally fed into a task decoder before generating the final outputs.

\par
\noindent
\textbf{Vision Encoder.} 
Since the vision transformer (ViT) \cite{dosovitskiy2020image} has shown great potential in vision representation extraction. Numerous research efforts \cite{kim2021vilt,li2021align, li2022blip, wang2022image} have introduced ViT or a variant form of it (such as Deit, CaiT and Swin Transformer) into VLP. In this paper, we focus on the original ViT \cite{dosovitskiy2020image} architecture since we only study the impact of model compression on performance. We build a series of vision encoders based on ViT with
different configurations, as shown in Table \ref{model_cfg}.

\par
\noindent
\textbf{Text Encoder.} 
The most widely used text encoder in VLP is BERT \cite{devlin2018bert}, which first segments the input sentence into a sequence of subwords, then insert two special tokens at the beginning and end of the sentence to generate the input text sequence. 
We also use the BERT as our default text encoder and build a series of text encoders based on BERT with
different configurations, as shown in Table \ref{model_cfg}.

\par
\noindent
\textbf{Multimodal Fusion.} 
There are two types of fusion, namely, merged attention and cross-attention. 
Previous work \cite{dou2022empirical} has proven that the cross-attention performs better than the merged attention model and computational efficiency. 
Therefore, we use cross-attention as our multimodal fusion operations, as shown in Figure \ref{framework}.
We control the size of the multimodal fusion module by setting different numbers of cross-attention layers.

\par
\noindent
\textbf{Task Decoder.}
The decoder module is optional in the VLP model. 
Many VLP models adopt the encoder-only architecture, where the cross-modal representations are directly fed into an output layer to generate the final outputs.
For transformer encoder-decoder architecture, the cross-modal representations are fed into a decoder and then to an output layer. 
More importantly, the encoder-decoder architecture is more flexible, as it can perform tasks such as image captioning, which may not be that straightforward for an encoder-only model to be applied to. Therefore, we take the decoder into account in our DLIP framework. 
The structure of the decoder is similar to the text encoder, and the difference is that it replaces the bidirectional self-attention layers with causal self-attention layers.

 There are many different model designs under the DLIP framework, in order to train a light and efficient model to better adapt to different downstream tasks, we use the multimodal mixture of encoder-decoder (MED) \cite{li2022blip} as our basic architecture, which can flexibly operate as a unimodal encoder or a multimodal encoder based on cross-attention, or a multimodal decoder.

\subsection{Pre-training Objectives}
For vision-language pre-training, we introduce three popular objectives: Image-Text Contrastive (ITC) Loss, Image-Text Matching (ITM) Loss and Language Modeling (LM) Loss, as described below.

\vspace{0.4em}
\par
\noindent
\textbf{Image-Text Contrastive Loss.}
Image-Text Contrastive (ITC) Loss aims to learn better unimodal representations and align the feature space of the visual encoder and the text encoder by encouraging positive image-text pairs to have similar representations in contrast to the negative pairs. 
The ITC loss has been demonstrated to be an effective objective for vision-language pre-training. 
We use the ITC loss following \cite{li2021align, li2022blip}, where a momentum encoder is introduced to produce features, and soft labels are created from the momentum encoder as training targets to account for the potential positives in the negative pairs.

\vspace{0.4em}
\par
\noindent
\textbf{Image-Text Matching Loss.}
Image-Text Matching (ITM) Loss aims to learn an image-text multimodal representation that captures the fine-grained alignment between vision and language. The multimodal fused representation predicts whether a pair of images and text is positive (matched) or negative (not matched). 
In order to find more informative negatives, we follow the hard negative mining strategy by  \cite{li2021align, li2022blip}, where negative pairs with higher contrastive similarity in a batch are more likely to be selected to compute the loss.

\vspace{0.4em}
\par
\noindent
\textbf{Language Modeling Loss.}
Language Modeling (LM) Loss aims to generate textual descriptions given an image. It optimizes a cross-entropy loss which trains the model to maximize the likelihood of the text in an autoregressive manner. 
Compared to the masked language modelling loss that has been widely-used for VLP, LM enables the model with the generalization capability to convert visual information into coherent captions.

The overall loss function of vision-language pre-training is:
\begin{equation}\label{vlp_loss}
\mathcal{L}_{\mathrm{VLP}}=\mathcal{L}_{\mathrm{ITC}}+\mathcal{L}_{\mathrm{ITM}}+\mathcal{L}_{\mathrm{LM}}
\end{equation}

\subsection{Distilling Objectives}
Now, we introduce how to distil VLP models. Figure \ref{framework} gives an overview of multimodal distillation.
We design multiple distillation objectives from hidden representation and attention information transfer perspectives.

\vspace{0.4em}
\par
\noindent
\textbf{Representation Distillation Loss.}
For hidden representation information transfer, previous work has demonstrated that the hidden representations aligning are an efficient distillation strategy to learn information efficiently from the teacher model \cite{jiao2019tinybert, fang2021compressing, wang2022efficientvlm}.
Therefore, we use the hidden representation distillation to minimize the divergence of the hidden representations. 
The objective is as follows: 
\begin{equation}\label{hr_loss}
    \mathcal{L}_{\mathrm{HR}}=\frac{1}{N} \sum_{i=1}^N \operatorname{cosineLoss} \left(\mathbf{H}_{i}^T \mathbf{W}_t, \mathbf{H}_{i}^S \mathbf{W}_s \right),
\end{equation}
where $N$ denotes the number of layers of the transformer. $\mathbf{W}_t$ and $\mathbf{W}_s$ is a learnable linear transformation that maps teacher and student representations into the same dimension space. We consider three kinds of hidden representation for the modal information transfer: vision representation, text representation and fusion representation. 

\vspace{0.4em}
\par
\noindent
\textbf{Attention-Based Distillation Loss.}
On the other hand, the attention mechanism has been a highly successful neural network component for NLP and CV tasks, which is also crucial for VLP.
The attention maps are computed via the following:
\begin{equation}
    \mathbf{A}=\operatorname{softmax}\left(\frac{\mathbf{Q} \mathbf{K}^{\top}}{\sqrt{d_k}}\right),
\end{equation}
where $Q$ and $K$ denote the query and key in the attention layer of a transformer block. $d_k$ is the dimension of the key as a scaling factor. 
Some works show that self-attention distributions of pre-trained LMs capture a rich hierarchy of linguistic information \cite{devlin2018bert, liu2019roberta}.
Transferring self-attention distributions has been used in previous works for Transformer distillation \cite{wang2020minilm, yang2022vitkd},
which optimizes the KL-divergence between the attention distributions of the teacher and student model:
\begin{equation}\label{eq_at}
    \mathcal{L}_{\mathrm{AT}}=\frac{1}{h|L|} \sum_{a=1}^{h} \sum_{l=1}^{|L|} D_{K L}\left(\mathbf{A}_{N, a, l}^T \| \mathbf{A}_{M, a, l}^S\right),
\end{equation}
where $L$ and $h$ represent the sequence length and the number of attention heads. $N$ and $M$ represent the number of layers for the teacher and student. $A^T_N$ and $A^S_M$ are the attention distributions of the $N$-th layer for the teacher and $M$-th layer for the student, respectively. We consider two kinds of attention-based maps for modal information transfer: self-attention and cross-attention maps.

Due to the differences in the architecture of the teacher and the student model, it is difficult to use the layer-to-layer strategy to transfer knowledge.
We chose the last transformer layer to transfer the knowledge as  \cite{wang2020minilm,fang2021compressing}, since distilling the last transformer block’s representation and attention allows more flexibility for the architecture of the student models and avoids the effort of finding the best layer mapping. 

Finally, we combined the pre-trained objectives with distilling objectives.
The overall objectives of DLIP are formulated as follows:
\begin{equation}
\mathcal{L}_{\mathrm{DLIP}}=  \mathcal{L}_{\mathrm{VLP}} +\mathcal{L}_{\mathrm{AT}} + \mathcal{L}_{\mathrm{HR}}.
\end{equation}

\begin{figure*}[t]
      \begin{center}
      \centerline{\includegraphics[width=175mm,height=64mm]{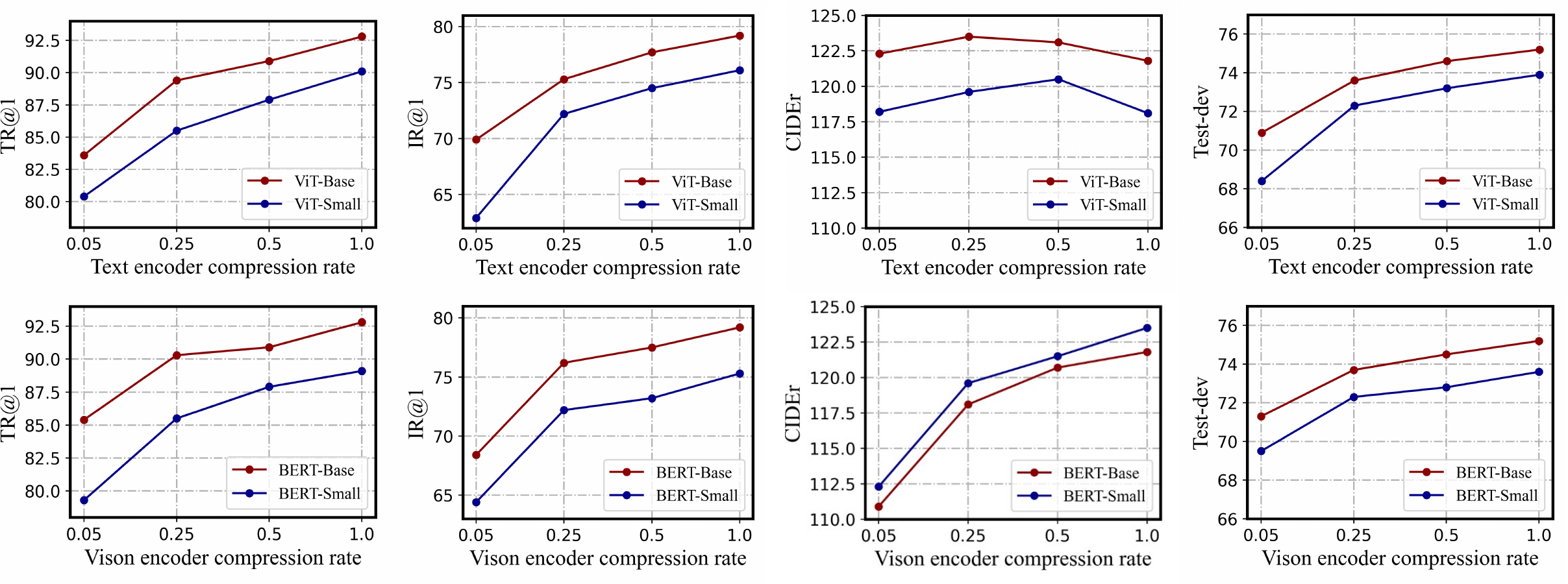}}
      \caption{
       The empirical study of the effects of different modal encoders under different compression ratios on various downstream tasks. }
      \vspace{-2.em}
      \label{model_size}
      \end{center}
      \vskip -0.1in
  \end{figure*}

\section{Experiments and Analyses}
In this section, we aim to ablate the role of different modules at different scales on the VLP model and the impact of different kinds of information in VLP distillation.
We conduct a series of controlled experiments on various VL tasks to derive practical insights for VLP Distilling.

\subsection{Experimental Settings}

\vspace{0.4em}
\par
\noindent
\textbf{Pre-training Datasets}
Following previous work \cite{li2021align,dou2022empirical}, we pre-train models on four commonly used datasets, including two human-annotated datasets: COCO \cite{lin2014microsoft} and Visual Genome \cite{krishna2017visual}. and two web datasets: Conceptual Captions \cite{sharma2018conceptual} and SBU Captions \cite{ordonez2011im2text}. 
The combined training data consists of about 4M images in total. 
We also experimented with an additional web dataset, Conceptual 12M \cite{changpinyo2021conceptual}, which contains 12M images with more noisy texts. 

\vspace{0.4em}
\par
\noindent
\textbf{Downstream Tasks.}
For ablation and analysis, we mainly focus on image-text Retrieval, image captioning  and visual question answering (VQA).  
For image-text Retrieval, we evaluate models for both image-to-text retrieval (TR) and text-to-image retrieval (IR) on COCO and Flickr30K.
For image captioning, all models were finetuned and evaluated on the COCO dataset. Following \cite{wang2021simvlm, li2022blip}, we add a prompt “a picture of” at the beginning of each caption, which leads to slightly better results.
For VQA, we follow \cite{li2021align,li2022blip} and consider it as an answer generation problem, where an image question is first encoded into multimodal embeddings and then given to an answer decoder.
We finetune the pre-trained model with both training and validation data in VQAv2 \cite{antol2015vqa}, and test the models on the test-dev set following the standard practice \cite{chen2020uniter}.

\vspace{0.4em}
\par
\noindent
\textbf{Implementation Details}
We pre-train our models using AdamW \cite{loshchilov2017decoupled} optimizer with a weight decay of 0.02. The learning rate is warmed-up to $5e-4$ and decayed linearly with a rate of 0.85. The image encoder is initialized from Vit pre-trained on ImageNet, and the text transformer is randomly initialized. We explore the effect of initialization on the model in a later section. We take random image crops of resolution 224$\times$224 during pre-training and increase the image resolution to 384$\times$384 during fine-tuning. By default, the model is trained on the 4M datasets.

\subsection{Impact of Model Compression}
To determine which modules could be compressed, we first perform an exploration study by comparing modal encoders of different sizes to determine the impact of compression of different modules on downstream tasks.
To this end, we design a series of models with different configurations, as shown in Table \ref{model_cfg}.
We pre-training with a combination of vision and text encoders of different scales and finetuning in various downstream tasks (\emph{e.g.,} VQA and image captions). 
The results are shown in Figure \ref{model_size}.
We can observe a lot of redundant parameters in the VLP model. 
Even though the model is compressed by 0.05$\times$, the performance does not drop as much as the compression ratio. 
We also observe that whether compressing an image encoder or a text encoder, the performance of most downstream tasks degrades, and the impacts of the different modal encoders are similar.
This means the image and text encoders are equally important for their performance on most VL tasks under the same compression scale.

We further experiment with the multimodal fusion module and control the module's size by setting different numbers of cross-attention layers. 
We test five values of the layer's number and show the results in Table \ref{abl_fusion}. 
As can be seen from the table, the model performance increases as the number of multimodal fusion layers increases, but when the number of multimodal fusion layers reaches a certain number ( $ l \geq 6 $ in our setting), there is no more significant improvement in the model performance, which means that the modal fusion is saturated. 

\begin{center}
\fcolorbox{black}{gray!10}{\parbox{.9\linewidth}{
\textbf{Takesways:} 
(\romannumeral1) Image and text encoders are equally important in model compression; 
(\romannumeral2) The large fusion module is unnecessary, and moderate fusion layers are beneficial and efficient.
}}
\end{center}

\begin{table}[t]
  \caption{The impact of Fusion modules.}
  \label{abl_fusion}
    \vspace{10.0pt}
  \begin{center}
  \begin{small}
  \renewcommand\arraystretch{1.2}
  \setlength{\tabcolsep}{2.2mm}
   \begin{tabular}{c|cccccc}
    \toprule 
    \multirow{2}{1.6cm}{\centering Fusion Layer}  &  \multirow{2}{0.7cm}{TR$@$1}  & \multirow{2}{0.7cm}{IR$@$1}  &  \multicolumn{1}{c}{Caption} & \multicolumn{1}{c}{VQA} \\
      &  & & CIDEr & test-dev   \\
    \midrule
    1     &  84.0 & 68.7 & 116.7 & 70.88  \\
    3     &  85.9 & 70.8 & 118.3 & 71.87  \\
    6     &  \textbf{88.0} & 72.7 & \textbf{119.1} & 72.65  \\
    9     &  87.3 & \underline{73.2} & 118.7 & \underline{72.74}  \\
    12    &  87.6 & \textbf{73.5} & \underline{118.9} & \textbf{73.00}  \\
  \bottomrule
  \end{tabular}
  \end{small}
  \end{center}
  \vskip -0.3in
\end{table}

\subsection{Importance of Different Information}
To investigate the impact of different information transfers on downstream tasks is also essential in the distillation, we evaluate the information transfer efficiency of hidden representation and attention maps.

For representation information, we compared the visual-textual fusion information and unimodal information. 
The knowledge of the teacher model is divided into 4 parts, visual information (\textbf{$\mathcal{L}_{Img}$}), text information (\textbf{$\mathcal{L}_{text}$}), visual-textual information fused by the encoder (\textbf{$\mathcal{L}_{VL_E}$}) and visual-textual information fused by the decoder (\textbf{$\mathcal{L}_{VL_D}$}).
We distil the student model using different information separately. 
Table \ref{abl_info} shows the results, which present different information all plays a positive role in distillation, but multimodal fusion information can improve the performance of student models better than unimodal information in distillation.
In addition, for the VLP model with a decoder, we observe that if only the encoders of the student model learn information from the teacher model through distilling, there is a limited improvement in downstream tasks that depend on the decoders. 
When the decoder of the student model is also learning through distilling, the performance of the student model can be greatly improved.

\begin{table}[t]
  \caption{The effect of different modal information.}
  \label{abl_info}
    \vspace{10.0pt}
  \begin{center}
  \begin{small}
  \renewcommand\arraystretch{1.2}
  \setlength{\tabcolsep}{1.5mm}
   \begin{tabular}{cccc|cccccc}
    \toprule 
    \multirow{2}{0.6cm}{\centering  $\mathcal{L}_{Img}$}  &  \multirow{2}{0.6cm}{\centering  $\mathcal{L}_{Text}$} &  \multirow{2}{0.6cm}{\centering  $\mathcal{L}_{VL_E}$} & \multirow{2}{0.7cm}{\centering  $\mathcal{L}_{VL_D}$} & \multirow{2}{0.7cm}{TR$@$1}  & \multirow{2}{0.7cm}{IR$@$1}  &  \multicolumn{1}{c}{Caption} & \multicolumn{1}{c}{VQA} \\
      &  &  & & & & CIDEr & test-dev   \\
    \midrule
    \XSolidBrush  &  \XSolidBrush &  \XSolidBrush  &  \XSolidBrush &   85.5  & 72.1 & 119.6 & 71.24         \\
    \CheckmarkBold  &  \XSolidBrush &  \XSolidBrush  &  \XSolidBrush &  86.9 & 73.1 & 120.2 & 72.37  \\
    \XSolidBrush  &  \CheckmarkBold &  \XSolidBrush  &  \XSolidBrush & 87.3  & 72.9 & 119.7 &  72.22 \\
    \CheckmarkBold  &  \CheckmarkBold &  \XSolidBrush  &  \XSolidBrush & 88.2  & 74.6 & 120.4 &  72.39 \\
    \XSolidBrush &  \XSolidBrush  &  \CheckmarkBold  & \XSolidBrush &  \underline{88.6} &  \underline{75.2} & 120.6 &  72.48 \\
    \XSolidBrush  &  \XSolidBrush   &  \XSolidBrush   & \CheckmarkBold &  87.9 & 74.5 &  \underline{121.3} &  \underline{72.61} \\
    \CheckmarkBold  &  \CheckmarkBold   &  \CheckmarkBold  & \CheckmarkBold  & \textbf{89.1} & \textbf{76.0}  & \textbf{121.9}  &  \textbf{72.70}  \\
    \bottomrule & 
    \end{tabular}
  \end{small}
  \end{center}
  \vskip -0.3in
\end{table}

\begin{table}[t]
  \caption{ \centering The effect of different attention information. }
  \label{abl_attn}
  \begin{center}
  \begin{small}
  \renewcommand\arraystretch{1.2}
  \setlength{\tabcolsep}{2.2mm}
   \begin{tabular}{cc|ccccc}
    \toprule 
    \multirow{1}{1.3cm}{\centering  Self-attn}  &  \multirow{1}{1.5cm}{\centering  Cross-attn} & \multirow{1}{0.7cm}{TR$@$1}  & \multirow{1}{0.7cm}{IR$@$1}  &  \multicolumn{1}{c}{Caption CIDEr} \\
    \midrule
    \XSolidBrush    &  \XSolidBrush     & 85.5 & 72.1 & 119.6   \\
    \CheckmarkBold  &  \XSolidBrush     & 87.2 & 72.9  & 120.4   \\
    \XSolidBrush    &  \CheckmarkBold     & \underline{87.7} & \underline{73.2}  & \underline{120.6}   \\
    \CheckmarkBold  &  \CheckmarkBold   & \textbf{88.2} & \textbf{73.6}  & \textbf{120.8}   \\
    \bottomrule
    \end{tabular}
  \end{small}
  \end{center}
  \vskip -0.2in
\end{table}

For attention information, we compared self-attention and cross-attention information. 
The results are shown in Table \ref{abl_attn}. 
The cross-attention performs better than self-attention in our setting, indicating
that it is more important to transfer multimodal information than unimodal information. 
This conclusion also applies to representation distillation.

Finally, we synthetically compare the effects of representation and attention-based distillation.
The main results are shown in Table \ref{abl_dist}. 
We can see that each distillation loss improves the overall performance, demonstrating the effectiveness of combing these objectives for pre-training.
In addition, we also notice that the representation information of the hidden layer can transfer information more efficiently than the attention map.

\begin{center}
\fcolorbox{black}{gray!10}{\parbox{.9\linewidth}{
\textbf{Takesways:} 
(\romannumeral1) The representation information is better than the attention information for distillation, and multimodal information is more efficient than unimodal information;
(\romannumeral2) For the VLP model with a decoder, the decoder requires separate distillation to improve the performance of decoder-based downstream tasks.
}}
\end{center}


\subsection{Impact of Initialization}
We find most works \cite{li2021align, dou2022empirical, li2022blip} initialize image and text encoders with different pre-trained ViTs \cite{dosovitskiy2020image} and BERT-like model \cite{liu2019roberta, devlin2018bert}. 
Initializing with a pre-trained model can be seen as a direct knowledge/information duplication.  
Therefore, we explore the impact of direct information inheritance of the unimodal model on VLP. 
We apply different initialization strategies to train the VLP model.
Vision-PT indicates that the image encoder is initialized with an image pre-trained model, and 
Text-PT indicates the text encoders are initialized with a text pre-trained model.
The results are shown in Table \ref{abl_pre}. There are no significant differences between the model performance of different text encoder initialization. However, the image pre-trained plays a vital role in VLP. 
It greatly affects the performance of the VLP model on downstream tasks.
Therefore, it is necessary to use a pre-trained model initialization both in the training and distillation of the VLP.

\begin{center}
\fcolorbox{black}{gray!10}{\parbox{.9\linewidth}{
\textbf{Takesways:} Initialization with pre-trained models is important for the visual encoder but has little impact on the text encoder. 
}}
\end{center}

\section{Distilling Language-Image Pre-training}
\label{DLIP}
In the above section, we separately evaluate the effect of different modules and models in the VLP training and distilling and derive practical insights. 
We now use these insights to distil a smaller, faster, lighter VLP model.

\begin{table}[t]
  \caption{The effect of different objectives.}
  \label{abl_dist}
  \vspace{10.0pt}
  \begin{center}
  \begin{small}
  \renewcommand\arraystretch{1.2}
  \setlength{\tabcolsep}{2.2mm}
   \begin{tabular}{ccc|cccccc}
    \toprule 
    \multirow{2}{0.6cm}{\centering  $\mathcal{L}_{VLP}$}  &  \multirow{2}{0.6cm}{\centering  $\mathcal{L}_{AT}$} &  \multirow{2}{0.6cm}{\centering  $\mathcal{L}_{HR}$} & \multirow{2}{0.7cm}{TR$@$1}  & \multirow{2}{0.7cm}{IR$@$1}  &  \multicolumn{1}{c}{Caption} & \multicolumn{1}{c}{VQA} \\
      &  &  & & & CIDEr & test-dev   \\
    \midrule
    \CheckmarkBold  &  \XSolidBrush &  \XSolidBrush  & 85.5  & 72.1 & 119.6 & 71.24         \\
    \CheckmarkBold &  \CheckmarkBold  &  \XSolidBrush   & 88.2 & 73.6  & 120.8 & 72.31   \\
    \CheckmarkBold  &  \XSolidBrush   &  \CheckmarkBold   & \underline{88.8} & \underline{75.4} & \underline{121.6} & \underline{72.62}   \\
    \XSolidBrush  &  \CheckmarkBold   &  \CheckmarkBold   &  87.4 & 74.8 & 120.9 & 72.26  \\
    \CheckmarkBold  &  \CheckmarkBold   &  \CheckmarkBold   &  \textbf{89.1} & \textbf{76.0}  & \textbf{121.9}  & \textbf{72.70}  \\
    \bottomrule
    \end{tabular}
  \end{small}
  \end{center}
  \vskip -0.1in
\end{table}

\begin{table}[t]
  \caption{\centering The impact of whether to use a pre-trained model. }
  \label{abl_pre}
  \vspace{5.0pt}
  \begin{center}
  \begin{small}
  \renewcommand\arraystretch{1.2}
  \setlength{\tabcolsep}{2.2mm}
   \begin{tabular}{cc|ccccc}
    \toprule 
    \multirow{1}{1.3cm}{\centering  Vision-PT}  &  \multirow{1}{1.3cm}{\centering  Text-PT} & \multirow{1}{0.7cm}{TR$@$1}  & \multirow{1}{0.7cm}{IR$@$1}  &  \multicolumn{1}{c}{Caption CIDEr} \\
    \midrule
    \XSolidBrush  &  \XSolidBrush  & 74.5 & 57.1 & 101.4           \\
    \CheckmarkBold &  \XSolidBrush     & \underline{85.5} & \textbf{72.1}  & 119.6     \\
    \XSolidBrush  &  \CheckmarkBold    & 75.9 & 58.5 & 100.5  \\
    \CheckmarkBold  &  \CheckmarkBold  & \textbf{85.9} & \underline{71.4}  & \textbf{119.8}     \\
    \bottomrule
    \end{tabular}
  \end{small}
  \end{center}
  \vskip -0.2in
\end{table}

  \begin{table*}[h]
    \caption{Comparisons with models pre-trained with different methods in model architecture, parameter size and Computational cost. DLIP distil from stronger BLIP\cite{li2022blip} and retains high accuracy on COCO caption and VQA task under different evaluating metrics using smaller parameters with faster inference. Our DLIP shows competitive results compared to other methods.}
    \label{main_result}
        \begin{small}
    \begin{center}
    \renewcommand\arraystretch{1.2}
    \setlength\tabcolsep{4.5pt}
    \begin{tabular}{c|ccccccccccc}
        \toprule
        \multirow{2}{2.0cm}{\centering Model}  & \multirow{2}{2.0cm}{ \centering Image encoder} & \multirow{2}{1.8cm}{ \centering Text encoder} & \multirow{2}{1.2cm}{\centering Params} & \multirow{2}{1.2cm}{\centering FLOPs} & \multirow{2}{1.2cm}{\centering Inference} & \multicolumn{2}{c}{COCO Caption} & \multicolumn{2}{c}{VQA v2}  \\
        &  &  &  &  & & B$@$4 & CIDEr  &  test-dev & test-std  \\
        \midrule
        OSCAR$_{\text{Base}}$ \cite{li2020oscar}     & R101-F &  BERT$_{\text{Base}}$ & 198.1M & 775.2G & 135.2ms & 36.5 & 123.7 & 73.2 & 73.4 \\
        BLIP$_{\text{Base}}$ \cite{li2022blip}     &  \multirow{2}{1.5cm}{ \centering ViT$_{\text{Base}}$} &  \multirow{2}{1.5cm}{ \centering BERT$_{\text{Base}}$} &  \multirow{2}{1.5cm}{ \centering 213.6M} &  \multirow{2}{1.5cm}{ \centering 17.8G} &  \multirow{2}{1.5cm}{ \centering 41.3ms} & 36.8 & 121.8  & 75.3 & 75.4 \\
        BLIP$_{\text{Base}}$ w/ 12M   & & & &  &  & 37.7 & 126.5  & 76.3 & 76.5 
          \\
        \midrule
        BLIP$_{\text{Mid}}$       & \multirow{3}{1.5cm}{ \centering ViT$_{\text{Middle}}$}  & \multirow{3}{1.5cm}{ \centering BERT$_{\text{Middle}}$}  & \multirow{3}{1.2cm}{ \centering 107.9M} & \multirow{3}{1.0cm}{ \centering 10.4G } & \multirow{3}{1.0cm}{ \centering 26.2ms} & 36.4 & 121.3  & 74.9 & 75.1 \\
        DLIP$_{\text{Mid}}$     &  &  &  &  &  & 36.9 & 122.4  & 75.5 & 75.6 \\
        DLIP$_{\text{Mid}}$ w/ 12M   &  &  &  &  &  & \textbf{37.3} & \textbf{125.7}  & \textbf{75.8} & \textbf{75.8} \\
        \midrule
        MiniVLM \cite{wang2020minivlm}  &  \multirow{2}{1.5cm}{ \centering TEE-0} &  \multirow{2}{1.5cm}{ \centering  MiniLM} &  \multirow{2}{1.5cm}{ \centering 53.2M} &  \multirow{2}{1.5cm}{ \centering 6.7G} &  \multirow{2}{1.5cm}{ \centering 23.6ms} & 35.6 & 119.8 & 69.4 & 69.1  \\
        DistillVLM \cite{fang2021compressing}      &  &  &  &  &  & 35.6 & 120.8 & 69.8 & 69.6 \\
        \midrule
        BLIP$_{\text{Small}}$  &  \multirow{3}{1.5cm}{ \centering  ViT$_{\text{Small}}$} &  \multirow{3}{1.5cm}{ \centering  BERT$_{\text{Small}}$} &  \multirow{3}{1.5cm}{ \centering 47.7M} &  \multirow{3}{1.5cm}{ \centering 4.4G}  &  \multirow{3}{1.5cm}{ \centering 14.9ms} & 36.2 & 119.6 & 71.2 & 70.9 \\
        DLIP$_{\text{Small}}$                 &  &  &  &  &  & 36.6 & 121.9 & 72.7 & 72.8 \\
        DLIP$_{\text{Small}}$ w/ 12M               & & &  &  &  & \textbf{37.2} & \textbf{124.7} & \textbf{73.1} & \textbf{73.4} \\
        \bottomrule 
    \end{tabular}
    \end{center}
     \end{small}
    \vskip -0.1in
  \end{table*}

  \begin{table}[h]
    \caption{Comparisons with models pre-trained with different methods under Flickr30k and COCO on image retrieval (IR) and text retrieval (TR) tasks in the finetuning setting. 12M means to use an additional Conceptual 12M dataset.}
    \vspace{-5.0pt}
    \label{ret_result}
    \begin{center}
    \renewcommand\arraystretch{1.2}
    \setlength\tabcolsep{4.5pt}
    \begin{tabular}{c|cc|cc}
        \toprule
        \multirow{2}{1.5cm}{\centering Model}  & \multicolumn{2}{c}{Flickr30k}   &  \multicolumn{2}{c}{COCO} \\
        & TR$@$1 & IR$@$1 & TR$@$1 & IR$@$1   \\
        \toprule
        OSCAR$_{\text{Base}}$ & 88.4 & 75.7 & 70.0 & 54.1     \\
        BLIP$_{\text{Base}}$  & 92.0 & 78.5 & 72.4 & 55.4     \\
        BLIP$_{\text{Base}}$ w/ 12M  & 83.3 & 75.6 & 57.9      \\
        \midrule
        BLIP$_{\text{Mid}}$   & 89.4 & 75.9 & 68.9 & 52.5   \\
        DLIP$_{\text{Mid}}$   & 92.7 & 79.6 & 73.1 & 55.3   \\
        \rowcolor{gray!10}  DLIP$_{\text{Mid}}$ w/ 12M & \textbf{58.2} & \textbf{82.3} & \textbf{76.2}	& \textbf{45.0}  \\
        \midrule   
        MiniVLM     & 81.1 &  68.5 & 58.8 &  43.9  \\
        DistillVLM  &  80.0 & 68.3 &  58.3 &  43.9   \\  
        \midrule
        BLIP$_{\text{Small}}$    & 85.5 & 72.1 & 64.9 & 49.4   \\
        DLIP$_{\text{Small}}$    & 89.1 & 76.0 & 68.8 & 52.6   \\
        \rowcolor{gray!10} DLIP$_{\text{Small}}$ w/ 12M & \textbf{91.5} & \textbf{79.3} & \textbf{71.5} & \textbf{55.1}  \\
        \bottomrule 
    \end{tabular}
    \end{center}
    \vskip -0.3in
  \end{table}

\begin{table}[t]
    \caption{The performance of different VLP using DLIP distillation. All models are trained with 4M image-text pairs.}
    \vspace{-5.0pt}
    \label{CLIP_result}
    \begin{center}
    \renewcommand\arraystretch{1.0}
    \setlength\tabcolsep{4.5pt}
    \begin{tabular}{c|cc|cc|c}
        \toprule
        \multirow{2}{1.5cm}{\centering Model}  & \multicolumn{2}{c|}{Flickr30k}  & \multicolumn{2}{c|}{COCO} &  \multicolumn{1}{c}{VOA} \\
         & TR$@$1 & IR$@$1 & TR$@$1 & IR$@$1 & test-dev    \\
        \toprule
        CLIP$_{\text{Base}}$   & 69.0 & 57.1 & 54.4 & 41.8 & -     \\
        CLIP$_{\text{Small}}$   & 30.4 & 25.2 & 33.8 & 26.7 & -     \\
        \rowcolor{gray!15}  DLIP$_{\textbf{Small}}$   & 34.1 & 30.1 & 36.4 & 28.6 & -     \\
        \midrule
        ALBEF $_{\text{Base}}$   & 93.4 & 79.4 & 71.8 & 54.9 & 74.1  \\
        ALBEF $_{\text{Small}}$  & 77.1 & 62.7 & 55.6 & 42.1 & 69.2   \\
        \rowcolor{gray!15}  DLIP $_{\textbf{Small}}$  & 80.3 & 64.2 & 58.2 & 44.3 & 70.5   \\
        \bottomrule 
    \end{tabular}
    \end{center}
    \vskip -0.2in
  \end{table}

\subsection{Main Results}
We compare the proposed DLIP with previous work, including MiniVLM \cite{wang2020minivlm} and DistillVLM \cite{fang2021compressing}. MiniVLM and DistilVLM are based on Oscar$_{\text{Base}}$ \cite{li2020oscar}. 
The results of our DLIP using BLIP$_{\text{Base}}$ \cite{li2022blip} as the teacher model. 
We choose two scale models as our student model: DLIP$_{Mid}$ and DLIP$_{Small}$.
The DLIP$_{Mid}$ uses ViT-Middle as image encoder and BERT-Middle as text encoder/decoder, respectively. The DLIP$_{Small}$ use ViT-Small as image encoder and BERT-Small as text encoder/decoder, respectively.
Detailed structural parameters refer to Table \ref{model_cfg}.
We use the non-distilled counterpart as our baselines.
\textbf{Note that} the reproduced BLIP models are trained from scratch and do not use the CapFilt \cite{li2022blip} for data cleaning. Therefore, the results shown in this paper are lower than those of the original BLIP \cite{li2022blip}.

As in Table \ref{main_result} and \ref{ret_result}, we list the larger transformer architectures VLP as the teacher model in the top group. Other groups are compressed models. 
From the results, we can see that our DLIP achieves substantial performance
improvement on multiple downstream VL tasks, and outperforms all compared baselines by a large margin despite DistilVLM and MiniVLM. Specifically, DLIP 
can achieve a VQA score of 73.4\% on the VQAv2 test-std set, and 124.7 in the CIDEr score on COCO Caption.
In particular, DLIP is able to compress BLIP\cite{li2022blip} by 1.9$\times$ from 213M to 108M parameters, achieving the same or better performance.
Moreover, it retains more than 95\% of the performance (TR@1:95.3 \emph{v.s} 91.5; IR@1:83.3 \emph{v.s} 79.3; CIDEr: 126.5 \emph{v.s} 124.7; VQA: 76.4 \emph{v.s} 73.4) on multiple tasks using 22.4\% parameters (213.6M \emph{v.s} 47.7M) and 24.8\% FLOPs (17.8G \emph{v.s} 4.4G) compared to the teacher model and accelerates its inference speed by 2.7$\times$ (41.3ms \emph{v.s} 14.9ms). 
We also draw the performance and efficiency trade-off diagram in Figure \ref{trades}.

\section{Conclusion} 
In this paper, We proposed DLIP, a simple yet effective distillation framework to train a light VLP model, through which we systematically explored how to compress the fully transformer-based vision-language pre-training through knowledge distillation efficiently.  
Empirical results show our DLIP achieves competitive performance after compressing the model substantially.

{\small
\bibliographystyle{ieee_fullname}
\bibliography{egbib}
}

\end{document}